\documentclass[fleqn,usenatbib]{mnras}

\usepackage{newtxtext,newtxmath}

\usepackage[T1]{fontenc}
\usepackage{threeparttable}
\usepackage{subfigure}
\usepackage{multirow}
\usepackage{booktabs}

\usepackage{xcolor}
\DeclareRobustCommand{\VAN}[3]{#2}
\let\VANthebibliography\thebibliography
\def\thebibliography{\DeclareRobustCommand{\VAN}[3]{##3}\VANthebibliography}

\usepackage{graphicx}	
\usepackage{amsmath}	

\title[Scientific Preparation for CSST]{Scientific Preparation for CSST: Classification of Galaxy and Nebula/Star Cluster Based on Deep Learning}

\author[Yuquan Zhang et al.]{
Yuquan Zhang,$^{1,2,3}$
Zhong Cao,$^{1,2,3,4}$\thanks{E-mail: zhongc@gzhu.edu.cn}
Feng Wang,$^{2,3,4}$\thanks{E-mail: fengwang@gzhu.edu.cn}
Lam, Man I,$^{5}$
Hui Deng,$^{2,3,4}$
Ying Mei,$^{2,3,4}$
and
Lei Tan,$^{2,3}$
\\
$^{1}$ School of Electronics and Communication Engineering, Guangzhou University, Guangzhou 510006, P.R. China\\
$^{2}$ Center For Astrophysics, Guangzhou University, Guangzhou 510006, P.R. China\\
$^{3}$ Great Bay Center, National Astronomical Data Center, Guangzhou, Guangdong, 510006, P.R. China\\
$^{4}$ Peng Cheng Laboratory, Shenzhen, 518000, P.R. China\\
$^{5}$ National Astronomical Observatories, Chinese Academy of Sciences, Beijing, 100101, P.R. China\\
}

\begin{document}
\label{firstpage}
\pagerange{\pageref{firstpage}--\pageref{lastpage}}
\maketitle

\begin{abstract}
The Chinese Space Station Telescope (abbreviated as CSST) is a future advanced space telescope. Real-time identification of galaxy and nebula/star cluster (abbreviated as NSC) images is of great value during CSST survey. While recent research on celestial object recognition has progressed, the rapid and efficient identification of high-resolution local celestial images remains challenging. In this study, we conducted galaxy and NSC image classification research using deep learning methods based on data from the Hubble Space Telescope. We built a Local Celestial Image Dataset and designed a deep learning model named HR-CelestialNet for classifying images of the galaxy and NSC. HR-CelestialNet achieved an accuracy of 89.09\% on the testing set, outperforming models such as AlexNet, VGGNet and ResNet, while demonstrating faster recognition speeds. Furthermore, we investigated the factors influencing CSST image quality and evaluated the generalization ability of HR-CelestialNet on the blurry image dataset, demonstrating its robustness to low image quality. The proposed method can enable real-time identification of celestial images during CSST survey mission.
\end{abstract}

\begin{keywords}
methods: data analysis – techniques: image processing
\end{keywords}



\section{Introduction} \label{1}
The Chinese Space Station Telescope (CSST) is a new space observatory designed for large-scale astronomical surveys. It is set to be launched into orbit by 2024, equipped with advanced technology and capabilities that are expected to significantly advance modern astronomy\citep{Zhan2021}.

To enhance the effectiveness of sky surveying, accurate and efficient identification of celestial bodies captured by the CSST survey camera in real-time is required in specific scenarios. Galaxies, nebulae/star clusters (NSCs) are extended source celestial bodies. The classification of these bodies can be readily performed when they are fully visible in the field of view. Moreover, the use of nearby stars enables the inference of the locations of the celestial bodies(\citealt{Groth1986}). However, in situations where only partial views of these celestial bodies are visible within the field of view, and their precise locations cannot be determined, direct identification of their types becomes challenging. 
Two factors contribute to this challenge: firstly, the morphological similarities between the partial views of different types of celestial bodies, such as elliptical galaxies and globular clusters, or irregular galaxies and open clusters, or galaxies with arms and nebulae. Secondly, it is difficult to determine the precise location of the object from surrounding stars in partial views. Therefore, advanced techniques for automated image classification are needed to aid in the identification process.

Since the introduction of the Hubble Sequence for classifying galaxy morphologies by \cite{Hubble1926}, research on image classification for galaxies, nebulae, and star clusters has been an active area of interest in astronomy, driven by the goal of developing faster and more accurate classification models. In the era prior to the widespread adoption of Charge Coupled Device (CCD), \cite{Storrie-Lombardi1992} and \cite{Owens1996} conducted research on galaxy classification by defining 13 morphological features within images derived from plate materials. They proved the feasibility and superiority of machine learning in galaxy image classification by adopting artificial neural network (ANN) and decision tree (DT) algorithm, respectively. 

An increasing number of galaxy images are being captured, facilitating the exploration of additional galaxy features and the design of more galaxy classification algorithms in the CCD era. \cite{Naim1995} extracted various global parameters such as ellipticity, surface brightness, as well as detailed parameters including arm number and arm length from galaxy images, resulting in a total of 24 predefined morphological features, which were utilized to construct a classification model using ANN. Although the classification process of these models was fully automated, feature extraction still relied on manual extraction, which reduced the overall efficiency of classification. Scholars began to study the automatic feature extraction of images. Texture \citep{Yamauchi2005}, entropy and spirality \citep{Ferrari2015}, and other automatic extraction features have been proposed one after another. Subsequently, some scholars focued on the automatic feature extraction of galaxy images. \cite{Calleja2004} took a distinct approach by refraining from defining and directly extracting features. Instead, each pixel value of the cropped galaxy image is considered as a feature. Similarly, \cite{Shamir2009} applied a series of image transformations, such as Fourier transforms and Wavelet transforms, to galaxy images and subsequently extracted 94 image features. Clearly, an excessive number of features in a classification model can introduce complexity and impede the attainment of superior model performance. On this basis, supervised learning methods, such as Fisher score, are commonly used for feature selection \citep{Shamir2009}, and unsupervised learning methods, such as principal component analysis (PCA), are used for feature dimensionality reduction\citep{Naim1995, Calleja2004}. These methods use automatic feature extraction and selection to effectively improve classification accuracy and efficiency through machine learning algorithms. 

In recent years, ground-based telescope survey projects, such as the Sloan Digital Sky Survey (SDSS) \citep{York2000}, the Cosmic Assembly Near-Infrared Deep Extragalactic Legacy Survey (CANDELS) \citep{Grogin2011} and the Dark Energy Survey (DES) \citep{Abbott2018}, have succeeded in obtaining important datasets containing a large number of galaxy images. Simultaneously, space-based telescope projects, including the Legacy ExtraGalactic UV Survey (LEGUS) and the Physics at High Angular Resolution in Nearby Galaxies with Hubble Space Telescope project (PHANGS-HST), have contributed massive high-resolution images of galaxies and star clusters \citep{Adamo2017, Cook2019, Lee2022}. 

The real-time processing and analysis of data is particularly important in the face of the massive astronomical images. The powerful feature extraction capability of convolutional neural networks (CNN) has made it a mainstream fully automated approach for image classification \citep{LeCun2015}.  \cite{Dieleman2015} first applied CNN to galaxy classification on the Galaxy Zoo2 dataset\citep{Willett2013}, achieving better performance over previous methods. Since then, state-of-the-art deep learning models in computer vision, such as AlexNet\citep{krizhevsky2012}, VGGNet\citep{Simonyan2014}, and ResNet\citep{He2016}, have been employed for morphological classification of galaxies and star clusters \citep{Zhu2019, Wei2020, Whitmore2021}. In addition, some self-designed deep learning models were also developed to recognize specialized datasets and achieve superior performance. For example, GAMORNET (Galaxy Morphology Network) was developed for galaxy classification \citep{Ghosh2020}, while STARCNET (STAR Cluster classification Network) was specifically designed for star cluster classification \citep{P2021}.  

 Previous studies have laid a solid foundation for classifying galaxies and NSCs. Nevertheless, several challenges have been posed to models to guarantee real-time identification of these objects in CSST surveys: (a) The classification of high-resolution single-exposure CSST images for specific science targets requires models that can classify such images rapidly and efficiently; (b) The majority of CSST single-exposure images are localized images of astronomical objects, some even faint, requiring models with robust capabilities to recognize both local and global features.

The remaining sections of this paper are organized as follows. In Sec. \ref{2}, we describe data sources and preprocessing, as well as the datasets used subsequently. The proposed model HR-CelestialNet and compared model, and their hardware requirements, are presented in Sec. \ref{3}. In Sec. \ref{4}, we present the HR-CelestialNet performance and compare with other models.  There is a brief discussion in Sec. \ref{5} and we summarize our work in Sec. \ref{6}.

\section{Data and Preprocessing}\label{2}

\subsection{Data Source}\label{2.1} 

The CSST survey camera is equipped with 18 multi-band imaging detectors, encompassing 7 wavelength bands (NUV, u, g, r, i, z, y) ranging from 255 to 1000 $\mathrm{nm}$. Each detector has a field of view (FoV) measuring $11\times11$ $ \mathrm{arcmin}^{2}$, a scale of $9k\times9k$ pixels, and an average pixel size of 0.074 \citep{Zhan2021}. Currently, researchers have employed the Spectral Energy Distribution (SED) method to simulate images based on the specific characteristics of the CSST\citep{Cao2018,Zhou2022}. Given the current low resolution of simulated celestial images, it is difficult to accurately describe this issue. In contrast, the Advanced Camera for Surveys Wide Field Camera (ACS/WFC) instrument operates within the wavelength coverage of 370 to 1100 $\mathrm{nm}$, has an FoV size of $202\times202$ $\mathrm{arcsec}^{2}$, and an average pixel size of 0.05 $\mathrm{arcsec}$ \citep{ACS}. And the Wide Field Camera 3 Ultraviolet-Visible (WFC3/UVIS) instrument operates within the wavelength coverage of 200 to 1000 $\mathrm{nm}$, has an FoV size of $162\times162$ $\mathrm{arcsec}^{2}$, and an average pixel size of 0.04 $\mathrm{arcsec}$ \citep{WFC3}. Thus, data products obtained by the ACS/WFC and WFC3/UVIS instruments can effectively meet the requirements of the CSST survey camera in terms of wavelength range, image resolution, and average pixel size parameters.

We downloaded and selected 7813 images from the Mikulski Archive for Space Telescopes (MAST) as the data source for model training, testing, and validating, which include 48 galaxies and 23 NSCs. The images are all single-exposure uncalibrated images obtained from 4101 raw FITS files. The images in these raw FITS files represent the initial products in the CSST survey. As mentioned in Sec. \ref{1}, performing classification at the initial product level has higher efficiency and better reflects engineering value. 

\subsection{Preprocessing}\label{2.2}

\begin{figure*}
\begin{center}

\subfigure[ACS/WFC image preprocessing]{
        \centering
        \includegraphics[width=0.95\textwidth]{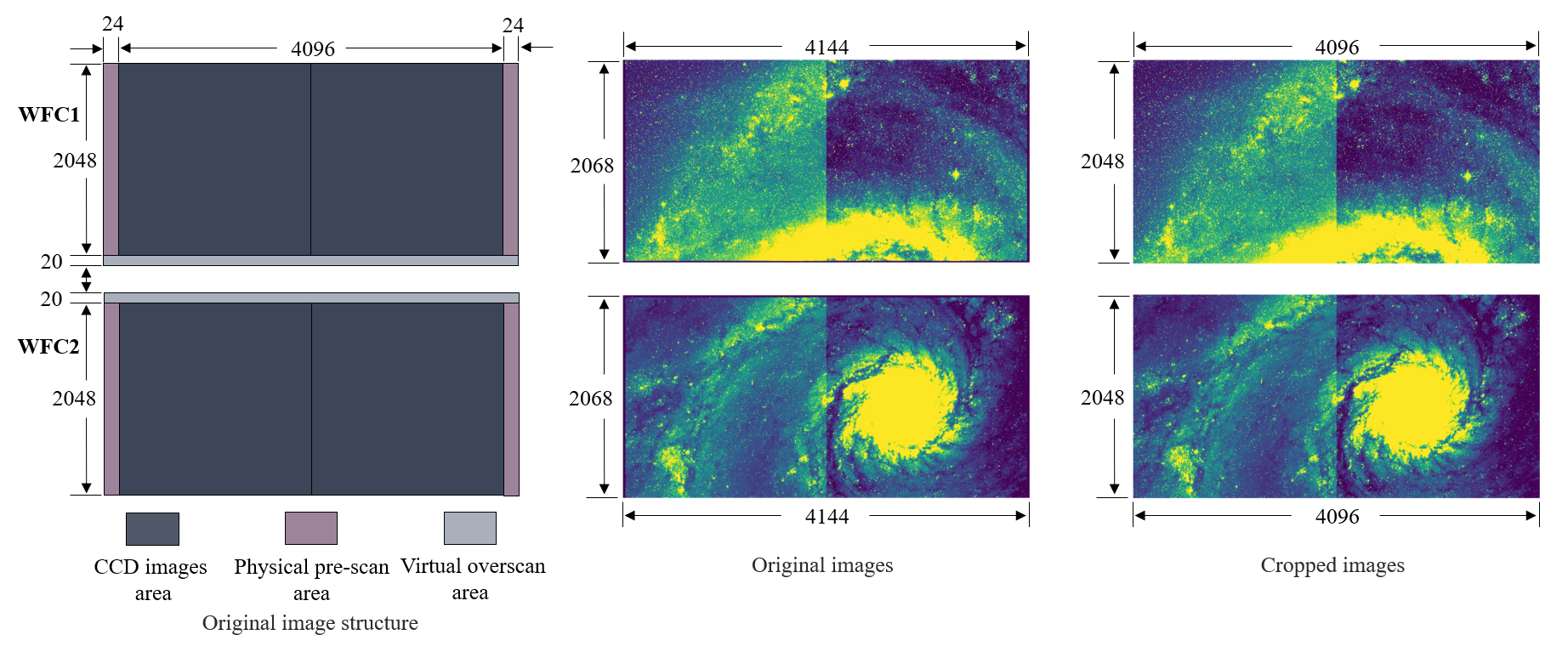}
        \label{acs_processing}
    }
\subfigure[WFC3/UVIS image preprocessing]{
	\includegraphics[width=0.95\textwidth]{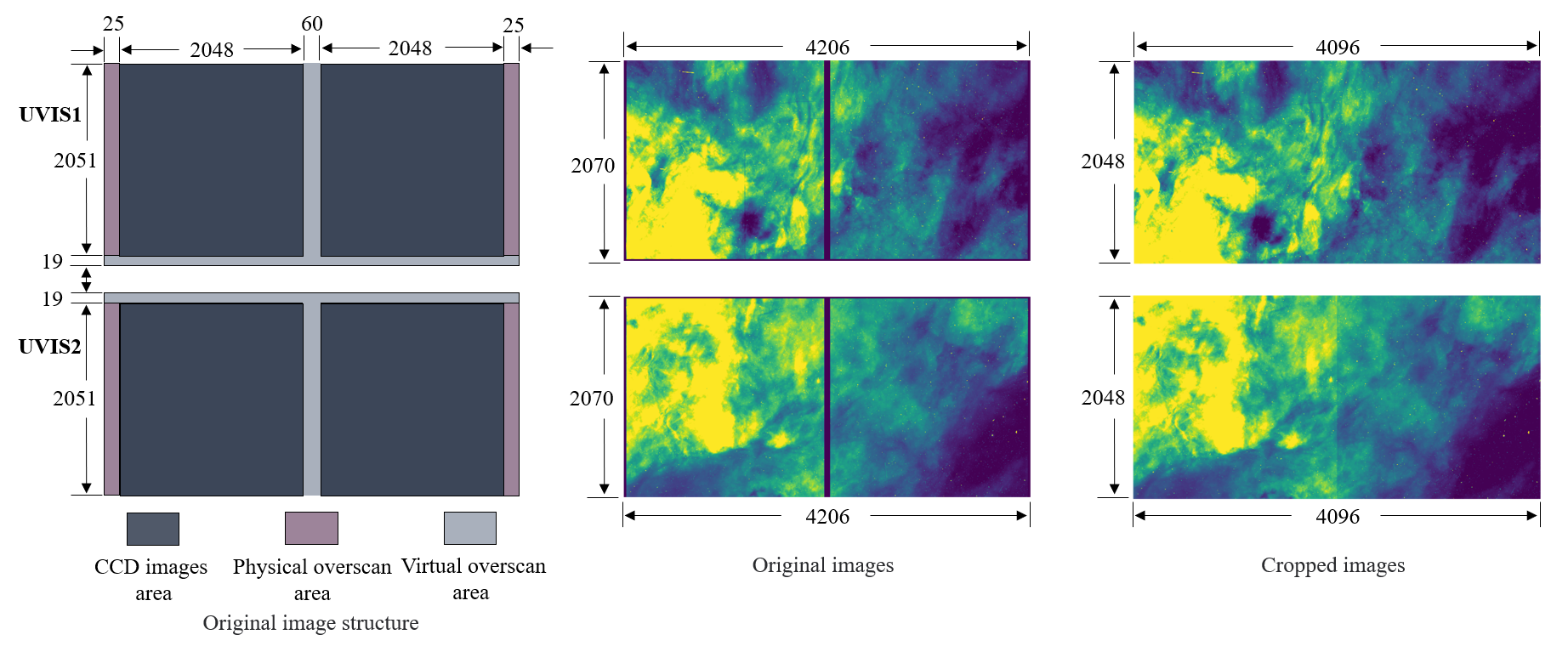}
	\label{wfc3_processing}
    }

\caption{Schematic diagram of the raw image preprocessing for ACS/WFC and WFC3/UVIS. (a) Left panel: ACS/WFC full-frame image layout. Middle panel: local raw image of the galaxy M51 taken by ACS/WFC. Right panel: corresponding cropped image samples. (b) Left panel: WFC3/UVIS full-frame image layout. Middle panel: local raw image of the Orion Nebula NGC1982 taken by WFC3/UVIS. Right panel: corresponding cropped image samples.}
\end{center}
\label{processing}
\end{figure*}

Both ACS/WFC and WFC3/UVIS are equipped with two CCDs. As each CCD corresponds to one image, each raw FITS file contains two corresponding image data units, resulting in two images per file. ACS/WFC's CCDs are named WFC1 and WFC2, and each CCD's image data unit consists of three parts, as shown in Fig.~\ref{acs_processing}. The first part is the CCD images area, which contains $2048\times4096$ pixels. The second part is the physical pre-scan area, located on both sides of the CCD images area, occupying 24 pixels, and used to estimate the bias level. The third part, the virtual overscan area, is situated at the bottom of the WFC1 CCD image area and the top of the WFC2 CCD image area, respectively, and has a horizontal width of 20 pixels. It serves the purpose of reading out the noise present in each single image. Therefore, the raw images corresponding to WFC1 and WFC2 all have $2068\times4144$ pixels.

Likewise, UVIS1 and UVIS2 are two CCDs equipped on WFC3/UVIS, as shown in Fig.~\ref{wfc3_processing}. The raw images of UVIS1 and UVIS2 consist of three parts:  the CCD images area, which comprises two regions of $2051\times2048$ pixels and contains a total of $2051\times4096$ pixels, with a horizontal width 3 pixels wider than that of WFC1 and WFC2; the physical overscan area, which is located on both sides of the CCD image area and occupies 25 pixels; the virtual overscan area in WFC3/UVIS differs from that of ACS/WFC, with two distinct components in its distribution. Horizontally, it is located at the bottom of the UVIS1 CCD image area and the top of the UVIS2 CCD image area, spanning 19 pixels. Vertically, it is situated between the two $2051\times2048$ regions of the CCD image area and spans 60 pixels. Therefore, the raw images corresponding to UVIS1 and UVIS2 consist of $2070\times4206$ pixels.

To obtain uniform image dimensions and retain only relevant CCD data, we applied specific preprocessing steps to the image units of both devices: (a) For both raw images of WFC1 and WFC2, we notice that by directly cropping the pre-scan and virtual overscan areas, we were able to obtain two CCD images with a size of $2048\times4096$. (b) For the raw images obtained from UVIS1 and UVIS2 detectors, we first cropped out both the physical overscan and virtual overscan areas. Following this, we merged the two CCD image regions from each detector to generate two CCD images of size $2051\times4096$. To achieve a consistent image size, we removed three pixels horizontally from the bottom of the UVIS1 raw image and from the top of the UVIS2 raw image. This yielded two CCD images with a uniform size of $2048\times4096$. (c) After obtaining CCD raw images with a size of $2048\times4096$, which serve as high-resolution local celestial image samples, manual selection is necessary to determine whether these samples contain celestial structures. Therefore, not all images in the raw FITS file are suitable for use as samples, and only one image may be selected or none. After manual screening, we obtained a final data set of 7813 high-resolution local celestial image samples with a size of $2048\times4096$.

\subsection{Dataset}\label{2.3}

The samples are re-labeled as galaxy or NSC based on their celestial bodies. For each celestial body in our dataset, multiple local celestial images are acquired, some of which have similarities due to being captured at the same location with different exposure times or filters. To prevent data leakage, we partitioned the dataset based on celestial bodies rather than individual images. In other words, all images of a single celestial body are exclusively designated for either training or testing purposes. The dataset is subsequently randomly split into a training set and a testing set using an $8:2$ ratio, a widely adopted practice in astronomical image classification research \citep{Wei2020,P2021,Walmsley2022}. Subsequently, we obtained a properly partition dataset named the Local Celestial Image Dataset (LCID), as shown in Table~\ref{dataset}.

To appraise the recognition performance of our high-resolution local celestial image classification model, we generated a low-resolution dataset called LCID-Resize. CV2.resize\footnote{https://docs.opencv.org/4.x/da/d54/group\_\_imgproc\_\_transform.html} function in the Python OpenCV package is applied to resize each image from LCID to a fixed size of $224\times448$. 

\begin{table}
    \centering
    \caption{Celestial body and sample statistics of LCID.}
    \label{dataset}
    \begin{tabular}{lccc}
        \hline
        \multirow{2}{*}{Statistics} & \multicolumn{3}{c}{Categories} \\
        \cline{2-4}
        & Galaxies & NSC & Total \\
        \hline
        Celestial bodies & 48 & 23 & 71 \\
        Samples (Training) & 3,310 & 2,908 & 6,218 \\
        Samples (Testing) & 852 & 743 & 1,595 \\
        Samples (Total) & 4,162 & 3,651 & 7,813 \\
        \hline
    \end{tabular}
\end{table}

\begin{figure*}
\begin{center}
  \subfigure[Clear NGC2835]{
        \centering
        \includegraphics[width=0.22\textwidth]{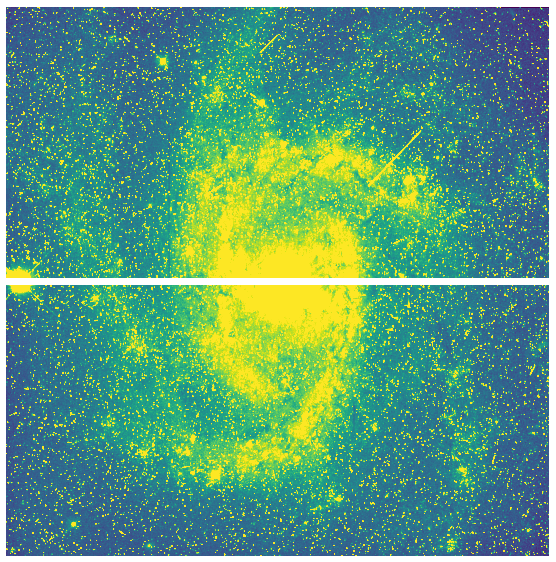}
    }
    \subfigure[Blurry NGC2835]{
	\includegraphics[width=0.22\textwidth]{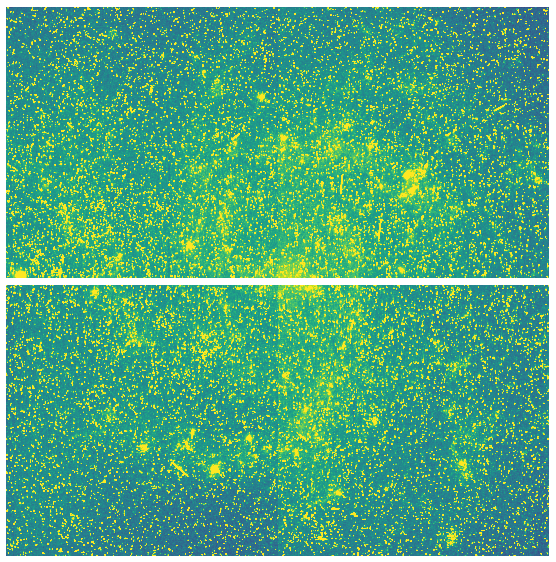}
    }
    \subfigure[Clear NGC1566]{
	\includegraphics[width=0.22\textwidth]{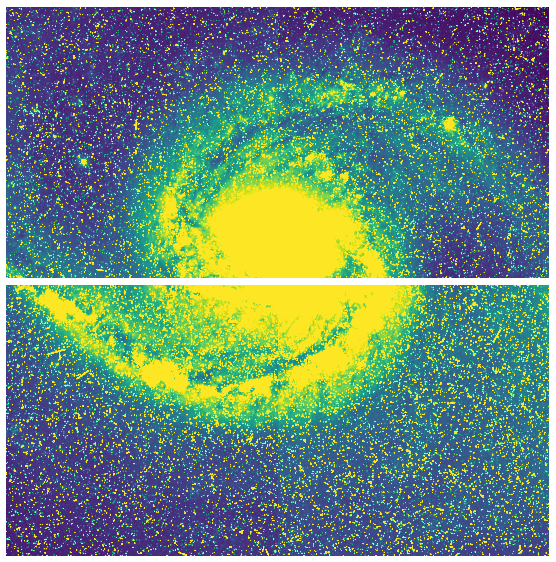}
    }
    \subfigure[Blurry NGC1566]{
	\includegraphics[width=0.22\textwidth]{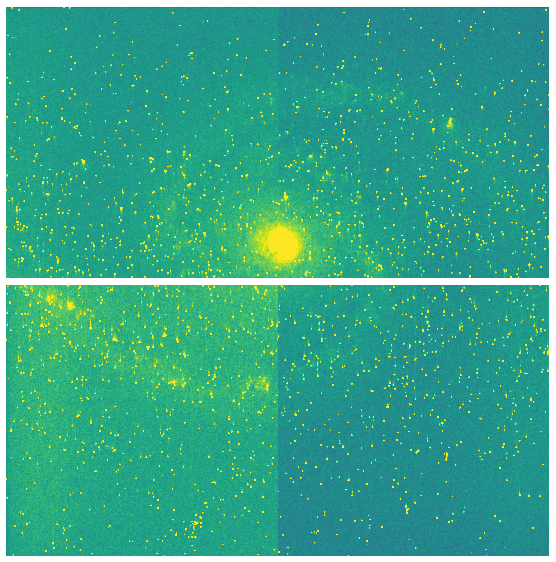}
    }
    \subfigure[Clear NGC6720]{
	\includegraphics[width=0.22\textwidth]{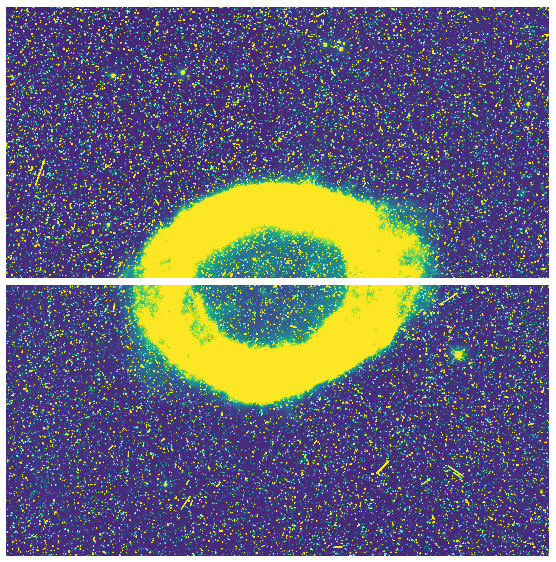}
    }
    \subfigure[Blurry NGC6720]{
	\includegraphics[width=0.22\textwidth]{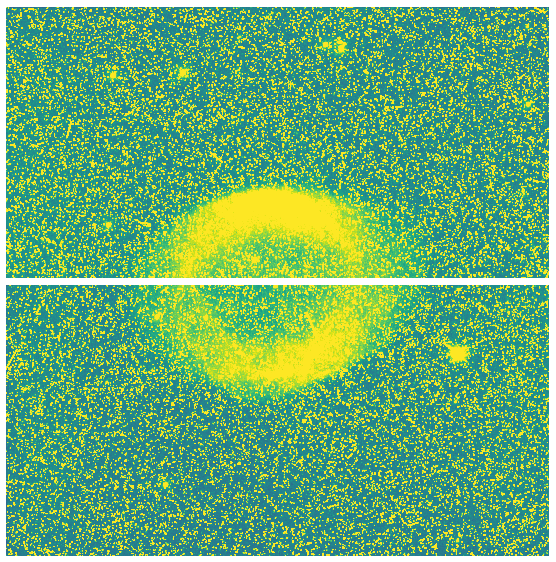}
    }
    \subfigure[Clear NGC1783]{
	\includegraphics[width=0.22\textwidth]{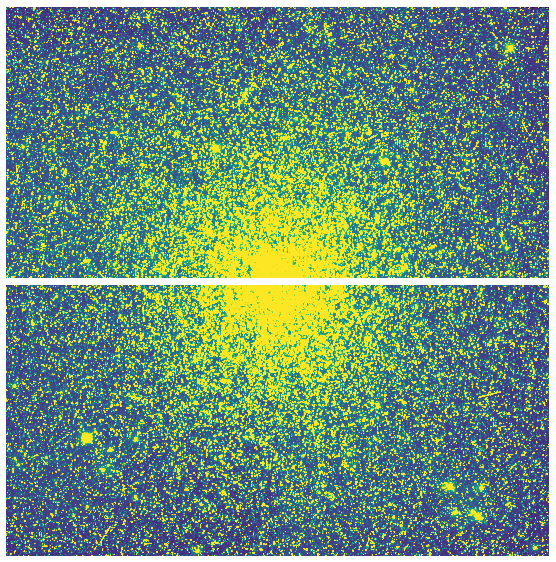}
    }
    \subfigure[Blurry NGC1783]{
	\includegraphics[width=0.22\textwidth]{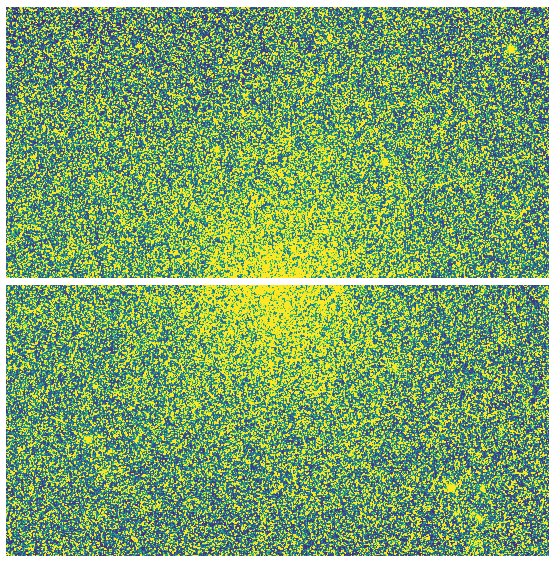}
    }
\caption{Clear and blurry samples in LCID.}
\label{fuzzy}
\end{center}
\end{figure*}

In the LCID dataset, some blurry samples are specially noticed, as illustrated in Fig.~\ref{fuzzy}. These samples exhibit fewer features compared to clear samples in the images. However, such samples are objectively present in telescopic observations. In the raw LCID image, the pixel values are obtained using a 16-bit analog-to-digital converter, with the unit expressed in data numbers. It is worth noting that in telescope observations, the image quality of celestial bodies is influenced by various factors. Although our image data is obtained through analog-to-digital conversion, with pixel units represented as Data Numbers, we can still express it in terms of electron counts (requiring only simple calibration procedures) \citep{ACS}. 
Therefore, in the raw image, under the assumption of neglecting noise, the theoretical flux of electron counts, denoted as $C$ can be described by equation~(\ref{ele}), where $t$ represents the exposure time, $A_{eff}$ denotes the effective aperture area of different filter telescopes, $S \left ( \lambda \right )$ represents the celestial SED, and $\tau \left ( \lambda \right ) = m_{eff} \left ( \lambda \right ) T_{filter} \left ( \lambda \right )$ represents the system throughput. Here, $m_{eff} \left ( \lambda \right )$ denotes the mirror efficiency, $T_{filter} \left ( \lambda \right )$ is the filter transmission, and $h$ and $c$ represent the Planck constant and the speed of light, respectively \citep{Cao2018}.

\begin{equation}
    C = t\ast A_{eff}\ast \int S \left ( \lambda  \right ) \tau \left ( \lambda  \right ) \frac{\lambda }{hc} \mathrm{d}\lambda  
	\label{ele}
\end{equation}

Therefore, the SED of the celestial body, exposure time, filter, CCD instrument effects are all factors that the image may become blurred. Accurate identification of such blurry samples will also be crucial in future CSST sky surveys. Accordingly, to evaluate the robustness of the models, a new validation dataset named LCID-Blurry was introduced, consisting exclusively of blurry samples. LCID-Blurry comprises 116 high-resolution celestial images, including 10 galaxies with 73 samples and 8 NSCs with 43 samples.

\section{Models}\label{3}
\subsection{HR-CelestialNet}\label{3.1}

We designed HR-CelestialNet, a deep learning model dedicated to directly recognizing high-resolution celestial images. Although state-of-the-art classification models focused on lower-resolution images (typically around $224\times224$ pixels), we strongly believed that high-resolution images contained a greater abundance of image features and compressing high-resolution images could result in the loss of important classification features.

HR-CelestialNet consists of 12 convolutional layers, 7 max-pooling layers, and 3 fully connected layers. The convolutional layers use kernel sizes of $7\times7$, $5\times5$, and $3\times3$, while the max-pooling layers employ pool kernels of size $8\times8$ with a stride of 4, $4\times4$ with a stride of 2, and $2\times2$ with a stride of 2. Notably, the number of channels in the convolutional layers gradually increases, starting with an output of 32 channels in the first layer and reaching a maximum output of 512 channels. After the convolutional and pooling layers, the feature map is flattened into a tensor of size $512\times3\times11$, and then passed through two fully connected layers, each consisting of 4096 neurons. To enhance the model's expressive power and prevent overfitting, a Rectified Linear Unit (ReLU) activation layer \citep{Relu} and a Batch Normalization (BatchNorm) layer \citep{BN} are applied after each convolutional layer. Furthermore, a ReLU activation function is inserted between the fully connected layers, and a Softmax function \citep{Softmax} is used to calculate and output the prediction weights for both classes. The complete architecture and parameter settings of the model are presented in Table~\ref{model_design}.

In Table~\ref{model_design}, we can consider HR-CelestialNet as composed of two segments based on the variation in output size. In the first 7 layers, the output size transforms from $1\times2048\times4096$ to $128\times121\times248$, hence, we refer to this segment as the "large-size learning component". From the eighth layer through the twenty-second layer, the output size progresses from $128\times121\times248$ to the final prediction output, which we designate as the "small-size learning component".

When designing the large-size learning component, we considered employing larger convolutional kernel sizes to accommodate larger feature map sizes and fewer convolutional layers. This strategy aimed to enhance the receptive field of this segment \citep{Luo2017}. Nevertheless, it's crucial to emphasize that larger convolutional kernels may not invariably lead to enhanced accuracy \citep{Zheng2020}. Therefore, based on empirical evidence and experimental results, we selected convolutional kernel sizes of $7\times7$ and $5\times5$. Besides, the selection of pooling kernel sizes and stride values for the pooling layers was guided by the downsampling ratio.

Regarding the small-size learning component, we drew inspiration from the design principles of VGGNet \citep{Simonyan2014} by incorporating their backbone network architecture. However, we adopted a global strategy of avoiding padding throughout the network. Experimental results have confirmed that this approach effectively reduces model parameters while preserving classification performance.

Based on the parameter size of HR-CelestialNet, a batch size of 4 was adopted. The cross-entropy loss function \citep{Cross} was employed to calculate the loss value for each mini-batch. The model parameters were optimized through the SGD optimizer \citep{SGD} with a learning rate of 0.0001. The training process ran for a total of 20 epochs. 

\begin{table*}
\begin{center}
\caption{Model design of HR-CelestialNet.}
\label{model_design}
\begin{tabular}{ccccccc}
\hline
Layer & Type & Kernel & Stride & Output Size & Activation & Trainable Paras\\
\midrule
- & Input & - & - & $1\times2048\times4096$ & - & 0\\
1 & Convolution & $7\times7$ & $1\times1$ & $32\times2042\times4090$ & ReLU & 1,600\\
2 & Max Pooling & $8\times8$ & $4\times4$ & $32\times509\times1021$ & - & 0\\
3 & Convolution & $7\times7$ & $1\times1$ & $64\times503\times1015$ & ReLU & 100,416\\
4 & Max Pooling & $4\times4$ & $2\times2$ & $64\times250\times506$ & - & 0\\
5 & Convolution & $5\times5$ & $1\times1$ & $128\times246\times502$ & ReLU & 204,928\\
6 & Convolution & $5\times5$ & $1\times1$ & $128\times242\times498$ & ReLU & 409,728\\
7 & Max Pooling & $2\times2$ & $2\times2$ & $128\times121\times248$ & - & 0\\
8 & Convolution &  $ 3\times3 $  & $1\times1$ & $256\times119\times247$ & ReLU & 295,168\\
9 & Convolution &  $ 3\times3 $  & $1\times1$ & $256\times117\times245$ & ReLU & 590,080\\
10 & Max Pooling & $2\times2$ & $2\times2$ & $256\times58\times122$ & - & 0\\
11 & Convolution &  $ 3\times3 $  & $1\times1$ & $256\times56\times120$ & ReLU & 590,080\\
12 & Convolution &  $ 3\times3 $  & $1\times1$ & $256\times54\times118$ & ReLU & 590,080\\
13 & Max Pooling & $2\times2$ & $2\times2$ & $256\times27\times59$ & - & 0\\
14 & Convolution &  $ 3\times3 $  & $1\times1$ & $512\times25\times57$ & ReLU & 1,180,160\\
15 & Convolution &  $ 3\times3 $  & $1\times1$ & $512\times23\times55$ & ReLU & 2,359,808\\
16 & Max Pooling & $2\times2$ & $2\times2$ & $512\times11\times27$ & - & 0\\
17 & Convolution &  $ 3\times3 $  & $1\times1$ & $512\times9\times25$ & ReLU & 2,359,808\\
18 & Convolution &  $ 3\times3 $  & $1\times1$ & $512\times7\times23$ & ReLU & 2,359,808\\
19 & Max Pooling & $2\times2$ & $2\times2$ & $512\times3\times11$ & - & 0\\
20 & Fully Connected & - & - & 4096 & ReLU & 69,210,112\\
21 & Fully Connected & - & - & 4096 & ReLU & 16,781,312\\
22 & Output & - & - & 2 & Softmax & 8,194\\
\hline
\end{tabular}

\begin{tablenotes}
\centering
            \item Note. The BatchNorm layer following each convolutional layer has been omitted in this table.
        \end{tablenotes}
\end{center}
\end{table*}

\subsection{Compared Models}\label{3.2}
We introduced some state-of-the-art classification models in the field of computer vision, namely AlexNet \citep{krizhevsky2012}, VGGNet \citep{Simonyan2014}, and ResNet \citep{He2016}, to compare the performances with HR-CelestitalNet. We conducted adaptive fine-tuning on these models based on the characteristics of the dataset to ensure the operation of the models.
Firstly, due to the single-channel nature of the LCID and LCID-Resize datasets mentioned in Sec. \ref{2}, these three models were modified to accept single-channel inputs. In addition, we observed that when recognizing high-resolution images, these models retained feature maps with large spatial dimensions before the fully connected layers, leading to increased model size and complexity. Therefore, when recognizing the LCID dataset, we introduced an Adaptive Average Pooling layer (AAPL) \citep{Adaptive} before the fully connected layers to maintain the model's size. The output size of the AAPL was consistent with the size of the feature maps before the fully connected layer when the model was performing recognition on the LCID-Resize dataset. Specifically, for AlexNet, the introduced AAPL produced an output size of $6\times13$, for VGGNet, the output size was $7\times14$, and for ResNet, the output size was $1\times2$.

\subsection{Hardware Requirements Comparison}\label{3.3}

Given the high-resolution image recognition capability of HR-CelestialNet and comparable models, they are acknowledged to be more computationally demanding than models designed for low-resolution images. While Table~\ref{model_design} details the trainable parameters for each layer of HR-CelestialNet, we also compared hardware requirements to provide more intuitive insight into model complexity and its correlation with performance.

We utilized the torchsummary.summary \footnote{https://pypi.org/project/torch-summary/} function from the torch-summary Python library to calculate the model size and the estimated total size required for training. As presented in Table~\ref{hard}, following fine-tuning, the sizes of AlexNet, VGGNet, and ResNet remained unchanged at 385.42 MB, 904.22 MB, and 42.62 MB, respectively. In comparison, our designed HR-CelestialNet exhibited a moderate size of 370.21 MB. When considering a batch size of 4 and recognizing images with a size of $224\times448$, the estimated total memory consumption for AlexNet, VGGNet, and ResNet was small, at 0.44 GB, 3.40 GB, and 0.53 GB, respectively. However, for image recognition tasks involving larger image sizes, specifically $2048\times4096$, the estimated total memory usage of AlexNet, VGGNet, and ResNet significantly increased to 6.21 GB, 211.14 GB, and 41.17 GB, respectively. In comparison, our HR-CelestialNet exhibited an estimated total memory usage of 32.79 GB, which fell within an acceptable range of hardware requirements.

\begin{table*}
\begin{center}
\caption{Comparison of hardware requirements for CNN models.}
\label{hard}
\renewcommand\tabcolsep{12pt}
\begin{tabular}{cccccc}
\hline
\multirow{2}{*}{Model} & \multirow{2}{*}{Data Set}  & Image Size & Input Size$^a$ & Model Size$^b$  & Estimated total Size$^c$ \\
&& (Pixels) &  (MB) & (MB) & (GB)\\
\hline
\multirow{2}{*}{AlexNet} & LCID-Resize & $224\times448$ & 1.53 & 385.42 & 0.44\\
 & LCID & $2048\times4096$ & 128 & 385.42 & 6.21\\
\hline
\multirow{2}{*}{VGGNet} & LCID-Resize & $224\times448$ & 1.53 & 904.22 & 3.40 \\
 & LCID & $2048\times4096$ & 128 & 904.22 & 211.14 \\
\hline
\multirow{2}{*}{ResNet} & LCID-Resize & $224\times448$ & 1.53 & 42.62 & 0.53 \\
 & LCID & $2048\times4096$ & 128 & 42.62 & 41.17\\
\hline
HR-CelestialNet & LCID   & $2048\times4096$ & 128 & 370.21 & 32.79 \\
\bottomrule
\end{tabular}
\end{center}
\begin{tablenotes}

            \item Note. a: The memory occupied by each batch of input samples, the batch size is 4, and each pixel in the image occupies 4 bytes; b: Model parameters occupy memory, as a visual description of model complexity; c: The total memory required by the estimated model during training, typically representing the GPU memory demand.
        \end{tablenotes}

\end{table*}

\section{Result}\label{4}

\subsection{Models Performance}\label{4.1}

In the field of computer vision, it is common to evaluate and compare model performance using metrics such as accuracy, precision, recall, and F1 score for binary classification tasks. Considering our scientific objective of identifying galaxy and NSC classes and considering them equally important, we focused on accuracy and F1 scores for both classes as our classification metrics.

Accuracy represents the straightforward measure of how well a model correctly classifies samples. F1 scores, on the other hand, provide a comprehensive and balanced evaluation by considering both precision and recall. To calculate these metrics, we first determined the number of true positives (TP), false positives (FP), true negatives (TN), and false negatives (FN) based on the predicted results and the ground truth labels of the test set. TP and TN denote the number of correctly predicted samples, while FP and FN represent the number of misclassified samples. Subsequently, we calculated the accuracy and F1 scores using equation~(\ref{acc}) to equation~(\ref{f1}).

\begin{equation}
    Accuracy = \frac{TP+TN}{TP+FN+TN+FP}
	\label{acc}
\end{equation} 

\begin{equation}
    Precision = \frac{TP}{TP+FP}
	\label{pre}
\end{equation}

\begin{equation}
    Recall = \frac{TP}{TP+FN}
	\label{recall}
\end{equation}

\begin{equation}
    F1 =\frac{2*Precision*Recall}{Precision+Recall}
	\label{f1}
\end{equation}

Table~\ref{model performance} presents the classification metrics for all models evaluated in our study.
On the LCID-Resize dataset, AlexNet showed an accuracy of 79.12\%, a galaxy F1 score of 78.75\%, and an NSC F1 score of 79.48\%, whereas VGGNet exhibited even better results with an accuracy of 88.28\%, a galaxy F1 score of 89.21\%, and an NSC F1 score of 87.17\%. Similarly, ResNet achieved an accuracy of 86.27\%, a galaxy F1 score of 87.05\%, and an NSC F1 score of 85.39\%. 

Comparing the models on the LCID dataset, AlexNet demonstrated an accuracy of 77.18\%, a galaxy F1 score of 76.58\%, and an NSC F1 score of 77.75\%. VGGNet achieved higher performance with an accuracy of 87.46\%, a galaxy F1 score of 88.15\%, and an NSC F1 score of 86.68\%, while ResNet obtained an accuracy of 87.65\%, a galaxy F1 score of 88.72\%, and an NSC F1 score of 86.35\%.
Our HR-CelestialNet achieved the best overall performance with an accuracy of 89.09\%, an F1 score of 90.20\% for galaxy, and an F1 score of 87.69\% for NSC on the LCID dataset. For the detailed classification metrics for all models on both datasets, please refer to Table~\ref{model performance}.

\begin{table*}
\centering
\caption{Classification metrics and time consumption.}
\label{model performance}
\renewcommand\tabcolsep{8pt}

\begin{tabular}{cccccccc}

\hline
\multirow{2}{*}{Model}& \multirow{2}{*}{Data Set} & \multicolumn{3}{c}{Classification Metrics} & \multicolumn{3}{c}{Time Consumption (ms/sample)} \\
\cline{3-8}& & Accuracy & F1(Galaxy) & F1(NSC) & Preprocessing & Classification & Total \\
\hline
\multirow{2}{*}{AlexNet} & LCID-Resize & 79.12\%  & 78.75\% & 79.48\% & 120.6 & 0.8 & 121.4 \\
& LCID & 77.18\% & 76.58\% & 77.75\% & 60.6 & 35.8 & 96.4 \\
    
\hline
 
\multirow{2}{*}{VGGNet} & LCID-Resize & 88.28\% & 89.21\% & 87.17\% & 120.6 & 4.7 & 125.3 \\
& LCID & 87.46\%  & 88.15\% &  86.68\% & 60.6 & 260.5 & 321.1 \\
    
\hline

\multirow{2}{*}{ResNet} & LCID-Resize & 86.27\% & 87.05\% & 85.39\% & 120.6 & 5.8 & 126.4 \\
& LCID & 87.65\% & 88.72\% & 86.35\% & 60.6 & 56.9 & 117.5 \\
\hline
HR-CelestialNet & LCID & 89.09\% & 90.20\% & 87.69\% & 60.6 & 55.9 & 116.5 \\
  \bottomrule
  \end{tabular}
\end{table*}

Additionally, we measured the time taken for each sample during the recognition process of these models on the same device. This involved measuring the average preprocessing time and average classification time to evaluate their recognition efficiency. The preprocessing time was dataset-dependent. For samples created from the LCID dataset following the steps outlined in Sec. \ref{2}, the average preprocessing time measured was 60.6 ms. For samples created from the LCID-Resize dataset, the average preprocessing time measured was 120.6 ms. 

The average classification time referred to the average time taken by the model to recognize a single sample. Among the models that recognize LCID, the lowest average classification time was observed for AlexNet(35.8 ms), followed by HR-CelestialNet (55.9 ms), ResNet (56.9 ms), and the highest time recorded for model VGGNet (260.5 ms). Moreover, among the models that recognize LCID-resize, the lowest average classification time was observed for AlexNet (0.8 ms), followed by VGGNet (4.7 ms), and ResNet (5.8 ms). 
Thus, we obtain the average total time required for each model to recognize a single sample. The fastest model was found to be AlexNet (LCID) with an average total time of 96.4 ms, followed by HR-CelestialNet with 116.5 ms, ResNet (LCID) with 117.5 ms, AlexNet (LCID-Resize) with 121.4 ms, VGGNet (LCID-Resize) with 125.3 ms, ResNet (LCID-Resize) with 126.4 ms, and VGGNet (LCID) with 321.1 ms.

In summary, HR-CelestialNet achieves state-of-the-art performance and efficiency for classifying high-resolution galaxy and NSC images, demonstrating its potential for practical astronomical image analysis applications. Overall, HR-CelestialNet strikes an optimal balance of effectiveness and efficiency.

\subsection{Robustness Analysis of Models on Blurry Samples}\label{4.2}
\begin{figure*}
\begin{center}
\includegraphics[width=0.98
\linewidth]{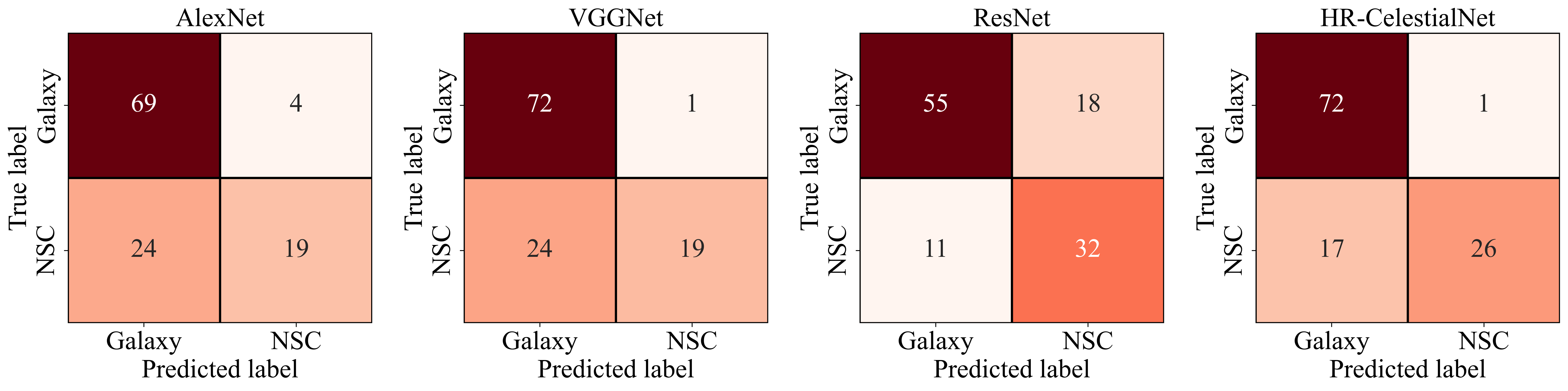}
\end{center}
\caption{Confusion matrices of models on LCID-Blurry validation set.}
\label{cm1}
\end{figure*}

As depicted in Fig.~\ref{cm1}, we utilized confusion matrices to offer a more intuitive depiction of the models' performance on blurry samples. In comparison with Table~\ref{model performance}, it is evident that the models' performance significantly declines on the LCID-Blurry validation set. Nevertheless, HR-CelestialNet remains the top performer with an accuracy of 84.48\%, followed by VGGNet with an accuracy of 78.45\%, AlexNet with an accuracy of 75.86\%, and finally ResNet with an accuracy of 75\%. Specifically, in the identification of category "galaxy"  within blurry samples, HR-CelestialNet, VGGNet, and AlexNet exhibit sustained high performance, with only 1 to 4 instances of misclassification. In contrast, ResNet demonstrates relatively poorer performance, misclassifying 18 samples. In the case of identifying category "NSC" within blurry samples, all models demonstrate moderate performance. Notably, ResNet excels with only 11 misclassifications, followed by HR-CelestialNet with 17 misclassifications, while VGGNet and AlexNet exhibit comparatively weaker performance, each with 24 misclassifications.

In summary, HR-CelestialNet showcases the optimal performance on blurry samples, demonstrating an approximate 6\% higher accuracy than the second-best performing model, VGGNet. However, further enhancements are required to improve the classification accuracy of blurry samples of NSC.

\begin{figure*}
  \centering
  \subfigure[Sample with inter-class similarity]{
        \centering
        \includegraphics[width=0.3\textwidth]{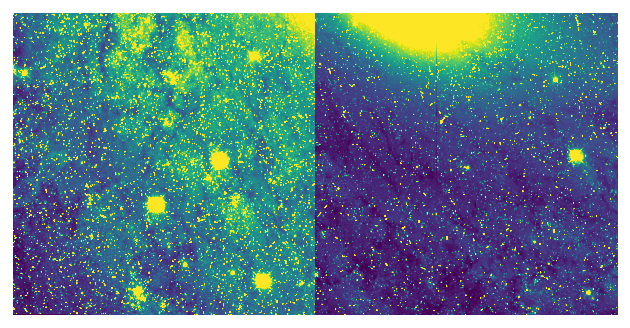}
        \label{mis_a}
    }
    \subfigure[Sample with inter-class similarity]{
	\includegraphics[width=0.3\textwidth]{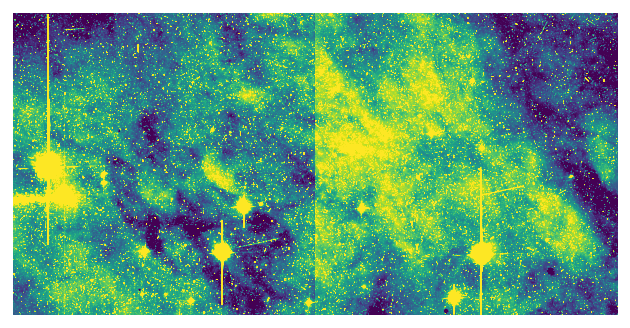}
	\label{mis_b}
    }
    \subfigure[Sample with insufficient features]{
	\includegraphics[width=0.3\textwidth]{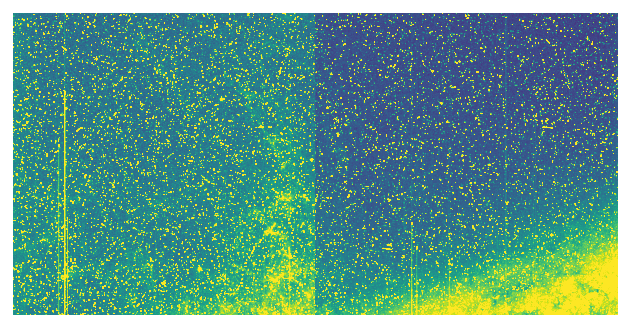}
	\label{mis_c}
    }
    \subfigure[Sample with insufficient features]{
	\includegraphics[width=0.3\textwidth]{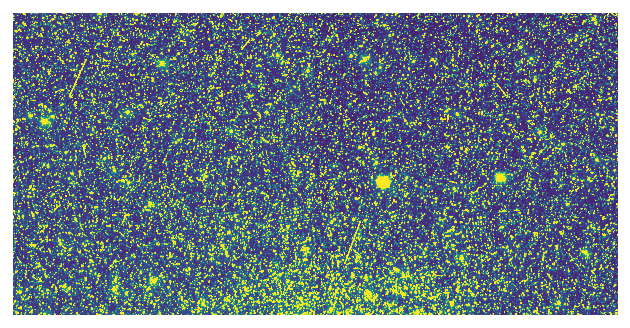}
	\label{mis_d}
    }
    \subfigure[Sample with significant noise]{
	\includegraphics[width=0.3\textwidth]{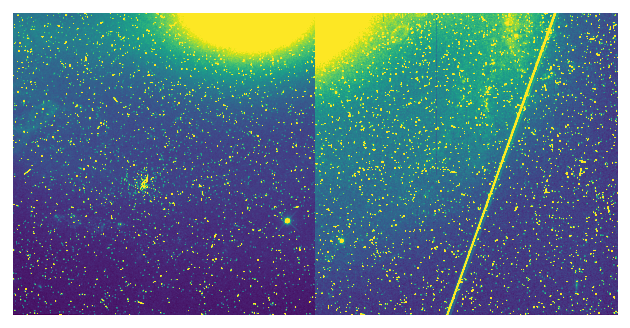}
	\label{mis_e}
    }
    \subfigure[Sample with significant noise]{
	\includegraphics[width=0.3\textwidth]{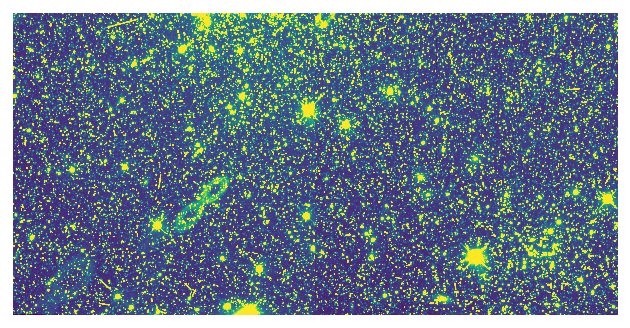}
	\label{mis_f}
    }
\caption{Misclassified samples from LCID.}
\label{mis}
\end{figure*}

\section{Discussions}\label{5}

\subsection{Model Advantage}\label{5.1}
Compared to the state-of-the-art deep learning models in the current computer vision domain, our designed HR-CelestialNet exhibits several advantages, including faster recognition speed and higher accuracy. Furthermore, experimental results on the LCID-Blurry validation set demonstrate the robustness of HR-CelestialNet in recognizing high-resolution celestial images.

In the design of HR-CelestialNet, we discarded padding and employed a combination of convolutional and pooling layers with various kernel sizes. This effectively reduces the dimension of feature maps, allowing the model to maintain a moderate number of parameters and hardware requirements while recognizing high-resolution images.
During the experiments, we categorized the data samples into two classes: galaxy and NSC, using local celestial images rather than complete celestial images. HR-CelestialNet achieved the highest accuracy of 89.09\%, with F1 scores of 90.20\% for galaxy and 87.69\% for NSC, and an average recognition time of 116.5 ms per sample. These results demonstrate HR-CelestialNet's superior performance in classifying local high-resolution celestial images, affirming its strong potential for various other celestial classification tasks.

\subsection{Misclassification}\label{5.2}
We analyzed misclassified samples. There are two types of misclassified samples in this study: galaxy samples misclassified as NSC, and NSC samples misclassified as galaxies. Analysis of these samples revealed three reasons for model misclassification: inter-class similarity, insufficient features, and image noise.

Inter-class similarity refers to the existence of similar features between galaxies and NSC samples. For instance, in Fig.~\ref{mis_a}, the sample is from the NGC6744 galaxy, where dense point sources resembling stars are present, a prominent feature of star cluster images. Similarly, in the Fig.~\ref{mis_b}, the sample is from M16, the Eagle Nebula, with a nebulous structure resembling the spiral arms of galaxies. The task of correct identification becomes challenging when typical features of another class are present in the samples.

Insufficient features occur when some samples in the dataset have a limited set of available features for classification. In Sec. \ref{2}, we specifically introduced the "Blurry Samples" in the dataset, primarily caused by the SEDs of celestial bodies, instrument effects of telescopes, and imaging parameters leading to a reduced number of features in the samples. However, in this context, we will discuss another scenario where the telescope only captures images of the edges of celestial objects. For instance, the Fig.~\ref{mis_c} represents a partial image of the M82 galaxy, capturing only a small portion of the galaxy. Similarly, the Fig.~\ref{mis_d} represents a partial image of the NGC 419 star cluster, showing only a small portion of the cluster. When the proportion of celestial body images in the sample is exceedingly limited or when the images are severely blurred, the model can only rely on these constrained features for classification, rendering such samples susceptible to misclassification.

Image noise in observed celestial images manifests as unwanted random signals or disturbances. Common types of noise in astronomical images include dark currents, thermal noise, and cosmic rays. Through the analysis of misclassified samples, we have found that cases of misclassification caused by these types of noise are rare, indirectly indicating that the model possesses a certain degree of anti-interference capability. However, there are exceptions. For instance, in the Fig.~\ref{mis_e}, a thick and bright slant line appears, which is likely caused by encountering a comet or the passage of an artificial satellite during the image acquisition process. Additionally, in Fig.~\ref{mis_f}, the appearance of a signal interference resembling the Arabic numeral "8" is highly likely to originate from the lens diffraction effect in the optical system. The presence of noise in astronomical images is inevitable, and the relatively low signal-to-noise ratio of primary image products remains a significant challenge for the model's anti-interference capability. 

In summary, we have identified three factors that can lead to misclassification of samples: inter-class similarity, insufficient features, and image noise. This indicates that our model requires further optimization, and simultaneously, the dataset needs to be supplemented with relevant subsets.

\section{Conclusions}\label{6}
In this study, according to the scientific design of CSST, we collected images of galaxies and NSCs from HST, and established three celestial body image datasets, including two high-resolution. Subsequently, we proposed a special classification model for galaxies and NSCs, HR-CelestialNet, and evaluated the model on AlexNet, VGGNet, and ResNet. The results show that HR-CelestialNet can identify single-channel, high-resolution celestial images, and will be able to perform the classification tasks of galaxies and NSCs in the CSST classification system in the future.  

In addition, we also discuss possible blurry samples in the future CSST. To explore this, we collected a dataset only containing blurry samples and re-performed the experiments. The results show that while the accuracy of all four models decreased slightly, HR-CelestialNet maintained a significantly higher ability to identify blurry samples compared to the models that identify low-resolution images.

In summary, HR-CelestialNet provides an effective solution for the classification of galaxies and NSCs in future CSST missions. Furthermore, the experimental results of HR-CelestialNet demonstrate its capability to directly process high-resolution images captured by CSST, such as learning celestial image features. This finding holds potential value for enabling in-depth research and exploration of celestial bodies in the future. In subsequent research, we will focus on balancing model performance and complexity, as well as improving the model's ability to accurately identify blurry samples, which will promote the development of the field of celestial image analysis and support the scientific mission of CSST.

\section*{Acknowledgements}
This work is supported by the National SKA Program of China (2020SKA0110300), the Funds for International Cooperation and Exchange of the National Natural Science Foundation of China (11961141001),
the National Science Foundation of China (12173028, 12373097), the Basic and Applied Basic Research Funds of Guangdong Province (2022A1515011558), the Fundamental and Application Research Project of Guangzhou (2023A03J0016), the Major Key Project of PCL.

\section*{Data Availability}
All the program code and part of the test data are stored in Github repository, the link is \url{https://github.com/YooQuuAN/Classification_of_galaxy_and_NSC.git}.

\bibliographystyle{mnras}
\bibliography{refer} 

\begin{thebibliography}{}
\makeatletter
\relax
\def\mn@urlcharsother{\let\do\@makeother \do\$\do\&\do\#\do\^\do\_\do\%\do\~}
\def\mn@doi{\begingroup\mn@urlcharsother \@ifnextchar [ {\mn@doi@}
  {\mn@doi@[]}}
\def\mn@doi@[#1]#2{\def\@tempa{#1}\ifx\@tempa\@empty \href
  {http://dx.doi.org/#2} {doi:#2}\else \href {http://dx.doi.org/#2} {#1}\fi
  \endgroup}
\def\mn@eprint#1#2{\mn@eprint@#1:#2::\@nil}
\def\mn@eprint@arXiv#1{\href {http://arxiv.org/abs/#1} {{\tt arXiv:#1}}}
\def\mn@eprint@dblp#1{\href {http://dblp.uni-trier.de/rec/bibtex/#1.xml}
  {dblp:#1}}
\def\mn@eprint@#1:#2:#3:#4\@nil{\def\@tempa {#1}\def\@tempb {#2}\def\@tempc
  {#3}\ifx \@tempc \@empty \let \@tempc \@tempb \let \@tempb \@tempa \fi \ifx
  \@tempb \@empty \def\@tempb {arXiv}\fi \@ifundefined
  {mn@eprint@\@tempb}{\@tempb:\@tempc}{\expandafter \expandafter \csname
  mn@eprint@\@tempb\endcsname \expandafter{\@tempc}}}

\bibitem[\protect\citeauthoryear{{Abbott} et~al.,}{{Abbott}
  et~al.}{2018}]{Abbott2018}
{Abbott} T.~M.~C.,  et~al., 2018, \mn@doi [\apjs] {10.3847/1538-4365/aae9f0},
  \href {https://ui.adsabs.harvard.edu/abs/2018ApJS..239...18A} {239, 18}

\bibitem[\protect\citeauthoryear{{Adamo} et~al.,}{{Adamo}
  et~al.}{2017}]{Adamo2017}
{Adamo} A.,  et~al., 2017, \mn@doi [\apj] {10.3847/1538-4357/aa7132}, \href
  {https://ui.adsabs.harvard.edu/abs/2017ApJ...841..131A} {841, 131}

\bibitem[\protect\citeauthoryear{{Cao} et~al.,}{{Cao} et~al.}{2018}]{Cao2018}
{Cao} Y.,  et~al., 2018, \mn@doi [\mnras] {10.1093/mnras/sty1980}, \href
  {https://ui.adsabs.harvard.edu/abs/2018MNRAS.480.2178C} {480, 2178}

\bibitem[\protect\citeauthoryear{{Cook} et~al.,}{{Cook}
  et~al.}{2019}]{Cook2019}
{Cook} D.~O.,  et~al., 2019, \mn@doi [\mnras] {10.1093/mnras/stz331}, \href
  {https://ui.adsabs.harvard.edu/abs/2019MNRAS.484.4897C} {484, 4897}

\bibitem[\protect\citeauthoryear{De~La~Calleja \& Fuentes}{De~La~Calleja \&
  Fuentes}{2004}]{Calleja2004}
De~La~Calleja J.,  Fuentes O.,  2004, \mn@doi [\mnras]
  {10.1111/j.1365-2966.2004.07442.x}, 349, 87

\bibitem[\protect\citeauthoryear{{Dieleman}, {Willett}  \& {Dambre}}{{Dieleman}
  et~al.}{2015}]{Dieleman2015}
{Dieleman} S.,  {Willett} K.~W.,   {Dambre} J.,  2015, \mn@doi [\mnras]
  {10.1093/mnras/stv632}, \href
  {https://ui.adsabs.harvard.edu/abs/2015MNRAS.450.1441D} {450, 1441}

\bibitem[\protect\citeauthoryear{{Dressel} \& {Marinelli}}{{Dressel} \&
  {Marinelli}}{2023}]{WFC3}
{Dressel} L.,  {Marinelli} M.,  2023, {WFC3 Instrument Handbook for Cycle 31 v.
  15.0}

\bibitem[\protect\citeauthoryear{{Ferrari}, {de Carvalho}  \&
  {Trevisan}}{{Ferrari} et~al.}{2015}]{Ferrari2015}
{Ferrari} F.,  {de Carvalho} R.~R.,   {Trevisan} M.,  2015, \mn@doi [\apj]
  {10.1088/0004-637X/814/1/55}, \href
  {https://ui.adsabs.harvard.edu/abs/2015ApJ...814...55F} {814, 55}

\bibitem[\protect\citeauthoryear{{Ghosh}, {Urry}, {Wang}, {Schawinski}, {Turp}
  \& {Powell}}{{Ghosh} et~al.}{2020}]{Ghosh2020}
{Ghosh} A.,  {Urry} C.~M.,  {Wang} Z.,  {Schawinski} K.,  {Turp} D.,   {Powell}
  M.~C.,  2020, \mn@doi [\apj] {10.3847/1538-4357/ab8a47}, \href
  {https://ui.adsabs.harvard.edu/abs/2020ApJ...895..112G} {895, 112}

\bibitem[\protect\citeauthoryear{{Grogin} et~al.,}{{Grogin}
  et~al.}{2011}]{Grogin2011}
{Grogin} N.~A.,  et~al., 2011, \mn@doi [\apjs] {10.1088/0067-0049/197/2/35},
  \href {https://ui.adsabs.harvard.edu/abs/2011ApJS..197...35G} {197, 35}

\bibitem[\protect\citeauthoryear{{Groth}}{{Groth}}{1986}]{Groth1986}
{Groth} E.~J.,  1986, \mn@doi [\aj] {10.1086/114099}, \href
  {https://ui.adsabs.harvard.edu/abs/1986AJ.....91.1244G} {91, 1244}

\bibitem[\protect\citeauthoryear{{Hardt}, {Recht}  \& {Singer}}{{Hardt}
  et~al.}{2015}]{SGD}
{Hardt} M.,  {Recht} B.,   {Singer} Y.,  2015, arXiv e-prints, \href
  {https://ui.adsabs.harvard.edu/abs/2015arXiv150901240H} {p. arXiv:1509.01240}

\bibitem[\protect\citeauthoryear{He, Zhang, Ren  \& Sun}{He
  et~al.}{2016}]{He2016}
He K.,  Zhang X.,  Ren S.,   Sun J.,  2016, in 2016 IEEE Conference on Computer
  Vision and Pattern Recognition (CVPR). p.~770, \mn@doi{10.1109/CVPR.2016.90}

\bibitem[\protect\citeauthoryear{{Hinton} \& {Salakhutdinov}}{{Hinton} \&
  {Salakhutdinov}}{2006}]{Cross}
{Hinton} G.~E.,  {Salakhutdinov} R.~R.,  2006, \mn@doi [Science]
  {10.1126/science.1127647}, \href
  {https://ui.adsabs.harvard.edu/abs/2006Sci...313..504H} {313, 504}

\bibitem[\protect\citeauthoryear{Hubble}{Hubble}{1926}]{Hubble1926}
Hubble E.~P.,  1926, \mn@doi [\apj] {10.1086/143018}, 64, 321

\bibitem[\protect\citeauthoryear{Ian, Yoshua  \& Aaron}{Ian
  et~al.}{2016}]{Softmax}
Ian G.,  Yoshua B.,   Aaron C.,  2016, Deep Learning (Adaptive Computation and
  Machine Learning Series)

\bibitem[\protect\citeauthoryear{{Ioffe} \& {Szegedy}}{{Ioffe} \&
  {Szegedy}}{2015}]{BN}
{Ioffe} S.,  {Szegedy} C.,  2015, \mn@doi [arXiv e-prints]
  {10.48550/arXiv.1502.03167}, \href
  {https://ui.adsabs.harvard.edu/abs/2015arXiv150203167I} {p. arXiv:1502.03167}

\bibitem[\protect\citeauthoryear{Krizhevsky, Sutskever  \& Hinton}{Krizhevsky
  et~al.}{2012}]{krizhevsky2012}
Krizhevsky A.,  Sutskever I.,   Hinton G.~E.,  2012, in Pereira F.,  et~al.,
  eds, Advances in Neural Information Processing Systems 25. Curran Associates,
  Red Hook, NY, p.~1097

\bibitem[\protect\citeauthoryear{{LeCun}, {Bengio}  \& {Hinton}}{{LeCun}
  et~al.}{2015}]{LeCun2015}
{LeCun} Y.,  {Bengio} Y.,   {Hinton} G.,  2015, \mn@doi [\nat]
  {10.1038/nature14539}, \href
  {https://ui.adsabs.harvard.edu/abs/2015Natur.521..436L} {521, 436}

\bibitem[\protect\citeauthoryear{{Lee} et~al.,}{{Lee} et~al.}{2022}]{Lee2022}
{Lee} J.~C.,  et~al., 2022, \mn@doi [\apjs] {10.3847/1538-4365/ac1fe5}, \href
  {https://ui.adsabs.harvard.edu/abs/2022ApJS..258...10L} {258, 10}

\bibitem[\protect\citeauthoryear{Lin, Chen  \& Yan}{Lin
  et~al.}{2013}]{Adaptive}
Lin M.,  Chen Q.,   Yan S.,  2013, arXiv preprint arXiv:1312.4400

\bibitem[\protect\citeauthoryear{{Luo}, {Li}, {Urtasun}  \& {Zemel}}{{Luo}
  et~al.}{2017}]{Luo2017}
{Luo} W.,  {Li} Y.,  {Urtasun} R.,   {Zemel} R.,  2017, \mn@doi [arXiv
  e-prints] {10.48550/arXiv.1701.04128}, \href
  {https://ui.adsabs.harvard.edu/abs/2017arXiv170104128L} {p. arXiv:1701.04128}

\bibitem[\protect\citeauthoryear{{Naim}, {Lahav}, {Sodre}  \&
  {Storrie-Lombardi}}{{Naim} et~al.}{1995}]{Naim1995}
{Naim} A.,  {Lahav} O.,  {Sodre} L. J.,   {Storrie-Lombardi} M.~C.,  1995,
  \mn@doi [\mnras] {10.1093/mnras/275.3.567}, \href
  {https://ui.adsabs.harvard.edu/abs/1995MNRAS.275..567N} {275, 567}

\bibitem[\protect\citeauthoryear{Nair \& Hinton}{Nair \& Hinton}{2010}]{Relu}
Nair V.,  Hinton G.~E.,  2010, in Proceedings of the 27th International
  Conference on International Conference on Machine Learning. ICML'10.
Omnipress, Madison, WI, USA, p. 807–814

\bibitem[\protect\citeauthoryear{{Owens}, {Griffiths}  \& {Ratnatunga}}{{Owens}
  et~al.}{1996}]{Owens1996}
{Owens} E.~A.,  {Griffiths} R.~E.,   {Ratnatunga} K.~U.,  1996, \mn@doi
  [\mnras] {10.1093/mnras/281.1.153}, \href
  {https://ui.adsabs.harvard.edu/abs/1996MNRAS.281..153O} {281, 153}

\bibitem[\protect\citeauthoryear{{P{\'e}rez}, {Messa}, {Calzetti}, {Maji},
  {Jung}, {Adamo}  \& {Sirressi}}{{P{\'e}rez} et~al.}{2021}]{P2021}
{P{\'e}rez} G.,  {Messa} M.,  {Calzetti} D.,  {Maji} S.,  {Jung} D.~E.,
  {Adamo} A.,   {Sirressi} M.,  2021, \mn@doi [\apj]
  {10.3847/1538-4357/abceba}, \href
  {https://ui.adsabs.harvard.edu/abs/2021ApJ...907..100P} {907, 100}

\bibitem[\protect\citeauthoryear{{Ryon}}{{Ryon}}{2023}]{ACS}
{Ryon} J.~E.,  2023, {ACS Instrument Handbook for Cycle 31 v. 22.0}

\bibitem[\protect\citeauthoryear{Shamir}{Shamir}{2009}]{Shamir2009}
Shamir L.,  2009, \mn@doi [\mnras] {10.1111/j.1365-2966.2009.15366.x}, 399,
  1367

\bibitem[\protect\citeauthoryear{{Simonyan} \& {Zisserman}}{{Simonyan} \&
  {Zisserman}}{2014}]{Simonyan2014}
{Simonyan} K.,  {Zisserman} A.,  2014, \mn@doi [arXiv e-prints]
  {10.48550/arXiv.1409.1556}, \href
  {https://ui.adsabs.harvard.edu/abs/2014arXiv1409.1556S} {p. arXiv:1409.1556}

\bibitem[\protect\citeauthoryear{{Storrie-Lombardi}, {Lahav}, {Sodre}  \&
  {Storrie-Lombardi}}{{Storrie-Lombardi} et~al.}{1992}]{Storrie-Lombardi1992}
{Storrie-Lombardi} M.~C.,  {Lahav} O.,  {Sodre} L. J.,   {Storrie-Lombardi}
  L.~J.,  1992, \mn@doi [\mnras] {10.1093/mnras/259.1.8P}, \href
  {https://ui.adsabs.harvard.edu/abs/1992MNRAS.259P...8S} {259, 8P}

\bibitem[\protect\citeauthoryear{{Walmsley} et~al.,}{{Walmsley}
  et~al.}{2022}]{Walmsley2022}
{Walmsley} M.,  et~al., 2022, \mn@doi [\mnras] {10.1093/mnras/stab2093}, \href
  {https://ui.adsabs.harvard.edu/abs/2022MNRAS.509.3966W} {509, 3966}

\bibitem[\protect\citeauthoryear{{Wei} et~al.,}{{Wei} et~al.}{2020}]{Wei2020}
{Wei} W.,  et~al., 2020, \mn@doi [\mnras] {10.1093/mnras/staa325}, \href
  {https://ui.adsabs.harvard.edu/abs/2020MNRAS.493.3178W} {493, 3178}

\bibitem[\protect\citeauthoryear{{Whitmore} et~al.,}{{Whitmore}
  et~al.}{2021}]{Whitmore2021}
{Whitmore} B.~C.,  et~al., 2021, \mn@doi [\mnras] {10.1093/mnras/stab2087},
  \href {https://ui.adsabs.harvard.edu/abs/2021MNRAS.506.5294W} {506, 5294}

\bibitem[\protect\citeauthoryear{{Willett} et~al.,}{{Willett}
  et~al.}{2013}]{Willett2013}
{Willett} K.~W.,  et~al., 2013, \mn@doi [\mnras] {10.1093/mnras/stt1458}, \href
  {https://ui.adsabs.harvard.edu/abs/2013MNRAS.435.2835W} {435, 2835}

\bibitem[\protect\citeauthoryear{{Yamauchi} et~al.,}{{Yamauchi}
  et~al.}{2005}]{Yamauchi2005}
{Yamauchi} C.,  et~al., 2005, \mn@doi [\aj] {10.1086/444416}, \href
  {https://ui.adsabs.harvard.edu/abs/2005AJ....130.1545Y} {130, 1545}

\bibitem[\protect\citeauthoryear{{York} et~al.,}{{York}
  et~al.}{2000}]{York2000}
{York} D.~G.,  et~al., 2000, \mn@doi [\aj] {10.1086/301513}, \href
  {https://ui.adsabs.harvard.edu/abs/2000AJ....120.1579Y} {120, 1579}

\bibitem[\protect\citeauthoryear{Zhan}{Zhan}{2021}]{Zhan2021}
Zhan H.,  2021, \mn@doi [Chinese Science Bulletin] {10.1360/TB-2021-0016}, 66,
  1290

\bibitem[\protect\citeauthoryear{{Zheng}, {Qiu}, {Luo}  \& {Li}}{{Zheng}
  et~al.}{2020}]{Zheng2020}
{Zheng} Z.-P.,  {Qiu} B.,  {Luo} A.~L.,   {Li} Y.-B.,  2020, \mn@doi [\pasp]
  {10.1088/1538-3873/ab5ed7}, \href
  {https://ui.adsabs.harvard.edu/abs/2020PASP..132b4504Z} {132, 024504}

\bibitem[\protect\citeauthoryear{{Zhou} et~al.,}{{Zhou}
  et~al.}{2022}]{Zhou2022}
{Zhou} X.,  et~al., 2022, \mn@doi [\mnras] {10.1093/mnras/stac786}, \href
  {https://ui.adsabs.harvard.edu/abs/2022MNRAS.512.4593Z} {512, 4593}

\bibitem[\protect\citeauthoryear{{Zhu}, {Dai}, {Bian}, {Chen}, {Chen}  \&
  {Hu}}{{Zhu} et~al.}{2019}]{Zhu2019}
{Zhu} X.-P.,  {Dai} J.-M.,  {Bian} C.-J.,  {Chen} Y.,  {Chen} S.,   {Hu} C.,
  2019, \mn@doi [\apss] {10.1007/s10509-019-3540-1}, \href
  {https://ui.adsabs.harvard.edu/abs/2019Ap&SS.364...55Z} {364, 55}

\makeatother
\end{thebibliography}
\appendix

\section{Some samples of LCID}

\begin{table*}
\begin{center}
\caption{LCID samples.}
\renewcommand\tabcolsep{18pt}
\begin{tabular}{@{}ccccccc@{}}
\hline \hline
ID & Instrument & OBSID & Filter & RA$^{\circ}$ & DEC$^{\circ}$ & HDU Index$^a$\\
\hline
NGC628 & ACS/WFC & j96r23b7q & F435W & 24.152604 & 15.769880 & 1 \\ 
NGC628 & ACS/WFC & j96r23b7q & F435W & 24.179654 & 15.782398 & 4 \\ 
NGC6302 & WFC3/UVIS & ie3o05ywq & F673N & 258.422912 & -37.091733 & 1 \\ 
NGC6302 & WFC3/UVIS & ie3o05ywq & F673N & 258.441272 & -37.109340 & 4 \\ 

M101 & ACS/WFC & jc1373esq & F555W & 210.893431 & 54.374057 & 1 \\ 
M101 & ACS/WFC & jc1373esq & F555W & 210.871072 & 54.348267 & 4 \\ 
RMC136 & ACS/WFC & jb6wd6evq & F814W & 84.805574 & -69.207164 & 1 \\ 
RMC136 & ACS/WFC & jb6wd6evq & F814W & 84.734141 & -69.193331 & 4 \\ 
IC1954 & WFC3/UVIS & idxr57q6q & F336W & 52.864111 & -51.900839 & 1 \\ 
IC1954 & WFC3/UVIS & idxr57q6q & F336W & 52.899666 & -51.907415 & 4 \\ 
M32 & ACS/WFC & jcnw11o5q & F814W & 10.621762 & 40.868951 & 4 \\ 

NGC1559 & WFC3/UVIS & idxr11e0q & F275W & 64.414259 & -62.800284 & 1 \\ 
NGC1559 & WFC3/UVIS & idxr11e0q & F275W & 64.409500 & -62.777487 & 4 \\ 
NGC1982 & ACS/WFC & j8dw03yeq & F502N & 83.861574 & -5.265002 & 1 \\ 
NGC1982 & ACS/WFC & j8dw03yeq & F502N & 83.833402 & -5.271876 & 4 \\ 
NGC6302 & WFC3/UVIS & iaco01e4q & F656N & 258.448209 & -37.111782 & 1 \\ 
NGC6302 & WFC3/UVIS & iaco01e4q & F656N & 258.424634 & -37.098706 & 4 \\ 

NGC1097 & ACS/WFC & jdxk38fhq & F814W & 41.576483 & -30.247906 & 1 \\ 
NGC1097 & ACS/WFC & jdxk38fhq & F814W & 41.563183 & -30.274417 & 4 \\ 
M82 & ACS/WFC & j9oe02wcq & F660N & 148.864287 & 69.687912 & 1 \\ 
... & ... & ... & ... & ... & ... & ... \\
\hline
\end{tabular}
\end{center}
\begin{tablenotes}

            \item Note. a: A FITS file can contain multiple Header-Data Units (HDUs), and the HDU index indicates the position of this image within the FITS file.
        \end{tablenotes}

\label{total_table}
\end{table*}


\label{lastpage}
\end{document}